\documentclass{article}

\usepackage[preprint]{corl_2026} % Uncomment for pre-prints (e.g., arxiv); This is like ``final'', but will remove the CORL footnote.

\usepackage{graphicx}
\usepackage{amsmath}     % 解决 \text{} 命令未定义问题
\usepackage{amssymb}     % 可选，用于更多数学符号
\usepackage{booktabs}
\usepackage{subcaption}
\usepackage{titlesec}
\titlespacing*{\subsection}{0pt}{4pt}{1pt}
\title{EaDex: A Cross-Embodiment Dexterous Manipulation Framework from Low-Cost Demonstrations}

\author{
  Qian Zhao$^{1,2}$ \quad Xin Tong$^{2}$ \quad Chengdong Wu$^{1,*}$ \quad Yang Yang$^{3,*}$ \quad Yingtian Li$^{2,*}$\\
  $^{1}$Faculty of Robot Science and Engineering, Northeastern University\\
  $^{2}$Shenzhen Institute of Advanced Technology, Chinese Academy of Sciences\\
  $^{3}$School of Automation, Nanjing University of Information Science and Technology\\
  \texttt{zhaoq5@mails.neu.edu.cn, yt.li@siat.ac.cn}\\
}

\begin{document}
\maketitle

%===============================================================================

\begin{abstract}
    Dexterous manipulation learning has long been hindered by the high costs of data and training, as pure reinforcement learning typically requires large-scale interactive exploration and imitation learning depends on high-quality demonstrations that are expensive to collect. To address this problem, we propose EaDex, a multi-embodiment dexterous manipulation learning framework under low-cost demonstration conditions, which enables rapid generation of demonstration data and consequently reduces training time for efficient dexterous manipulation. At the data level, EaDex captures human hand motions using only a single RGB-D camera and constructs structured demonstration data through MANO-based hand modeling, data normalization, and motion retargeting. At the learning level, we introduce a contact-reward-based dynamic demonstration annealing mechanism, which guides early-stage exploration under demonstration and gradually transitions to autonomous optimization with accumulating contact rewards. Using our custom dataset, we evaluate EaDex on three dexterous hands and three articulated object-opening tasks, covering nine cross-embodiment manipulation settings, achieving a 55.3\% relative improvement over the baseline without demonstration annealing. These results validate the effectiveness of the proposed low-cost demonstration pipeline and the dynamic demonstration annealing strategy for dexterous manipulation learning.
\end{abstract}

%===============================================================================
\begin{figure}[!htbp]
    \centering
    \includegraphics[width=\linewidth]{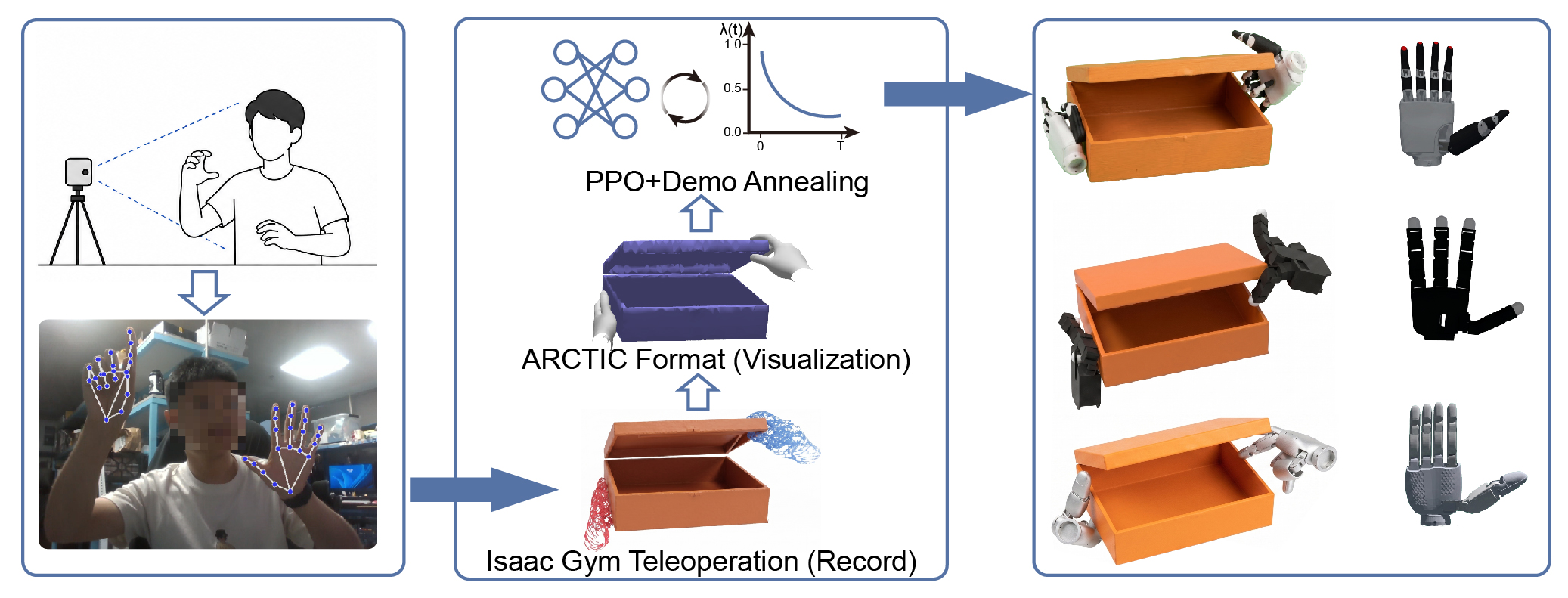}
    \caption{
    \textbf{Overview of EaDex}. 
    An end-to-end dexterous manipulation framework that captures human hand demonstrations using a single RGB-D camera and performs corresponding manipulation tasks across dexterous hands with different embodiments.
    }
    \label{fig1}
\end{figure}

\section{Introduction}	
Dexterous hands are important end-effectors in robotic systems, whose high degree-of-freedom structures enable fine-grained manipulation capabilities~\citep{ref1}. However, the resulting high-di\-mensional control space also substantially increases the difficulty of policy learning~\citep{ref2}. Existing methods mainly follow two technical paradigms: reinforcement learning (RL)~\citep{ref1,ref3,ref4,ref5} and imitation learning (IL)~\citep{ref6,ref7,ref8}. RL learns policies through trial-and-error interactions with the environment and can acquire generalizable behaviors in complex scenarios. Nevertheless, in dexterous manipulation, it often suffers from large exploration spaces, high training costs, and difficulty in convergence~\citep{ref9}. In contrast, IL bypasses the lengthy trial-and-error process by directly leveraging human or expert demonstrations, thereby improving sample efficiency~\citep{ref10,ref11}. However, its performance highly depends on the quality of demonstrations, and for dexterous hands, collecting high-precision demonstrations typically requires complex devices or expensive systems~\citep{ref12,ref13}, making it difficult to scale up. Recent works combine RL and IL by incorporating demonstrations into reinforcement learning rewards, which reduces exploration costs and improves adaptation to complex tasks, yet these methods still rely on costly demonstration datasets~\citep{ref14,ref15}. This contradiction motivates a practically important question: how can we stably learn dexterous manipulation policies from low-cost demonstration data~\citep{ref16}?

To address this problem, we study functional dexterous manipulation under low-cost and low-precision demonstration conditions. Our key observation is that, for articulated object manipulation, coarse-grained demonstrations can provide sufficient contact priors for effective policy learning, after which reinforcement learning can further optimize manipulation behaviors. Based on this observation, we propose \textbf{EaDex}, an end-to-end framework for dexterous manipulation learning. As shown in Fig.~\ref{fig1}, EaDex starts from human hand demonstrations captured by a single RGB-D camera and constructs a unified demonstration representation for multiple dexterous hand platforms through MANO-based hand modeling, ARCTIC-style data standardization, temporal smoothing, contact extraction, contact retargeting, and cross-hand motion retargeting. Building on this representation, EaDex trains policies using our custom demonstration dataset and introduces a dynamic demonstration annealing mechanism that gradually decreases the policy's reliance on the demonstrations as contact rewards accumulate.

We conduct systematic experiments on three dexterous hands with different embodiments, including the Inspire Hand, Allegro Hand, and XHand, and three articulated objects, including box, waffle iron, and mixer. Under low-cost demonstration conditions, EaDex achieves an average success rate of 36.5\% across nine cross-embodiment manipulation tasks, corresponding to a 55.3\% relative improvement over the baseline without demonstration annealing, and in some tasks, EaDex achieves a success rate of up to 93.3\%. These results demonstrate the effectiveness of the proposed framework in complex dexterous manipulation scenarios and its generalization ability across different dexterous hand embodiments.

\textbf{The main contributions of this work are summarized as follows:}

\begin{enumerate}
    \item We propose a low-cost demonstration acquisition and processing framework for dexterous manipulation. The framework leverages a single RGB-D camera to capture human hand demonstrations without relying on expensive teleoperation hardware or high-precision motion capture systems, and organizes the raw demonstrations into a unified structured representation suitable for policy learning.
    
    \item We introduce a contact-reward-based dynamic demonstration annealing mechanism. This mechanism adaptively adjusts the demonstration weight according to the policy's mastery of key contact structures, enabling a smooth transition from demonstration-guided learning to autonomous optimization.
    
    \item We validate the proposed method on three different dexterous hands and three articulated object tasks. The results demonstrate that our approach achieves scalable dexterous manipulation under low-cost demonstration conditions.
\end{enumerate}
	
%===============================================================================

\section{Related Work}
\label{sec:related}

\textbf{Reinforcement Learning for Dexterous Manipulation.}
RL has been widely applied to dexterous hand manipulation tasks, such as in-hand manipulation~\citep{ref1,ref17,ref18,ref19} and single-hand grasping~\citep{ref20,ref21,ref22}. However, due to the high degrees of freedom of dexterous hands, designing effective reward functions to guide policy exploration remains challenging~\citep{ref23}, especially in bimanual tasks where specifying task objectives or rewards that effectively guide exploration is difficult. Even with carefully designed rewards, pure RL typically requires a large number of simulated interactions to discover feasible contact patterns and manipulation sequences, leading to high computational cost. This motivates the use of demonstrations, which can serve both as an implicit task specification and as guidance on how to accomplish the task, thereby reducing ineffective exploration and training cost. Since running RL exploration directly on real hardware is costly~\citep{ref24}, simulation has become a common tool for training dexterous hand policies~\citep{ref10}. Therefore, this work also conducts policy learning in simulation and leverages demonstrations to guide RL, aiming to reduce the exploration and computational burden in high-dimensional dexterous manipulation. Compared with RL methods that rely on large-scale exploration~\citep{ref19}, this work focuses on using demonstration priors to constrain the policy search space, thereby improving the learning efficiency and success rate for dexterous manipulation of articulated objects~\citep{ref25}.

\textbf{Imitation Learning for Dexterous Manipulation.}
IL has been widely used to improve learning efficiency in dexterous manipulation by leveraging demonstration data. However, acquiring high-quality hand demonstrations typically requires complex teleoperation setups or precise calibration systems~\citep{ref26,ref27}, which makes it difficult to collect sufficient data to support generalization to unseen environments. Some approaches attempt to extract hand motions from videos~\citep{ref28,ref29}, but publicly available videos often lack the fidelity required to capture detailed hand states, as vision-based hand pose estimation is noisy and unreliable. \citet{ref30} proposed a single-camera teleoperation system, but it relies on carefully aligned customized hand models and is primarily limited to relatively simple tasks such as grasping. These limitations highlight the importance of low-cost acquisition of reliable demonstration data that can support more complex manipulation tasks.

\textbf{Algorithmic Progress and Demonstration-Guided Annealing.} 
Achieving human-level dexterity remains a core challenge in robotics~\citep{ref31,ref32}. Classical methods such as trajectory optimization~\citep{ref33,ref34} and STOCS~\citep{ref35} can compute joint trajectories and contact forces, but are mainly offline and thus limited for real-time applications. Model predictive control (MPC)~\citep{ref36} enables online planning over a finite time horizon, yet its computational burden hinders real-time deployment. Reinforcement learning (RL) optimizes policies through environmental interactions~\citep{ref19}, supporting complex task training, but the exploration space remains large. \citet{ref14} leverage human demonstrations to provide key guidance and rapidly focus policy search to address this challenge, while also introducing curriculum learning to prevent early failures. However, these methods rely on high-cost datasets~\citep{ref12} and cannot generalize to actions not seen in the dataset. \citet{ref15} similarly use human demonstrations to accelerate learning and combine dynamic PPO to improve success rates beyond pure imitation, yet they still depend on expensive datasets~\citep{ref13} and remain difficult to generalize to manipulation skills not covered by the dataset. Inspired by these works, we introduce demonstration-guided learning to reduce RL training cost, collect demonstration data in a low-cost manner to facilitate scaling to different scenarios, and incorporate demonstration-guided annealing to improve policy success rate.

%===============================================================================

\begin{figure}[!htbp]
    \centering
    \includegraphics[width=\linewidth]{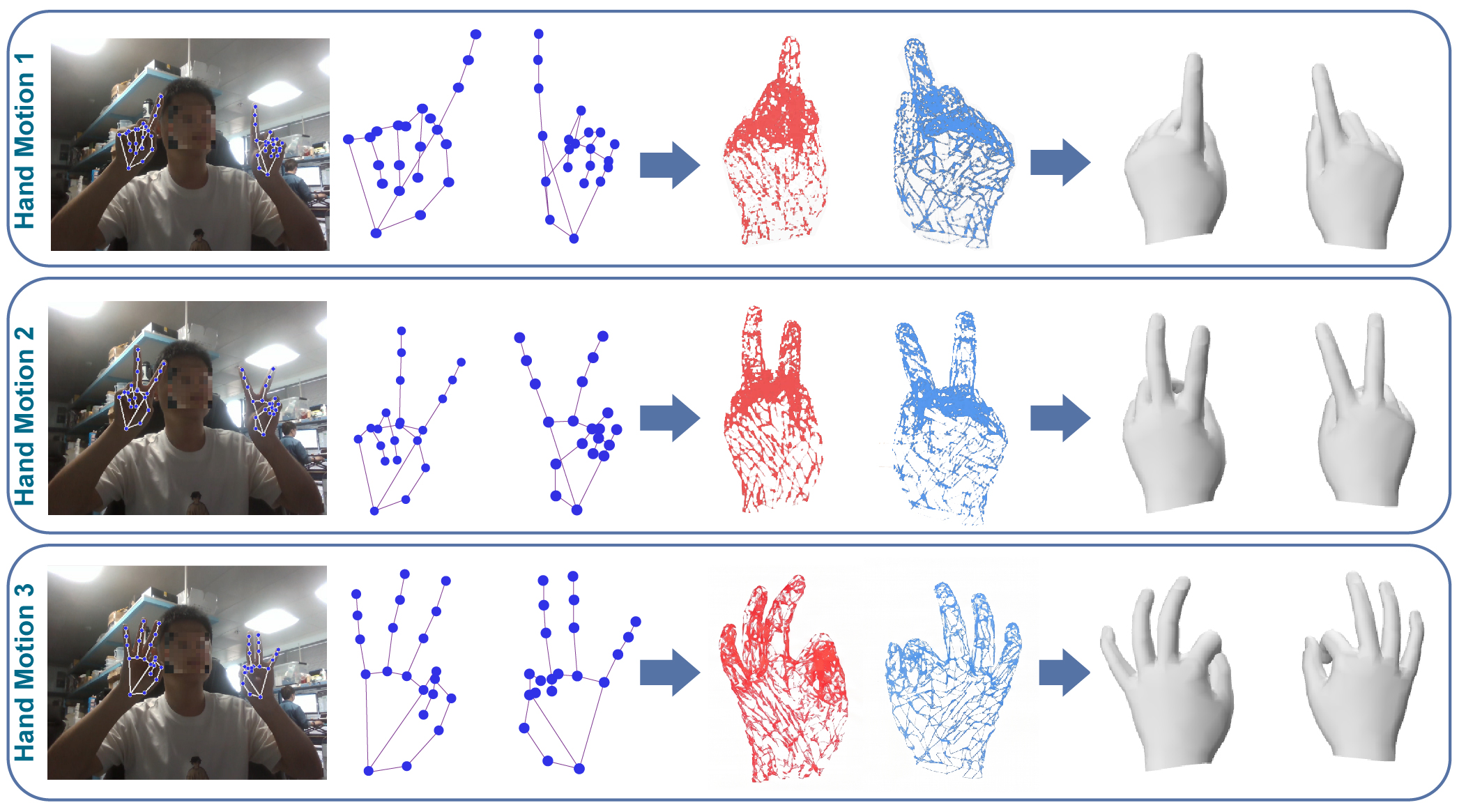}
    \caption{
    \textbf{Hand Gesture Demonstration Reconstruction}.
    EaDex enables precise replication of various hand gestures. 
    \textbf{Left:} Keypoints of human hands captured by the RGB-D camera in the real environment. 
    \textbf{Middle:} Hand models reconstructed in Isaac Gym from the captured keypoints. 
    \textbf{Right:} Visualization using the ARCTIC format viewer after saving the data.
    }
    \label{fig2}
\end{figure}

\section{Method}
\label{sec:method}

\textbf{Overview.} 
We propose EaDex, an end-to-end framework for dexterous manipulation. In Section~\ref{sec:data_collection}, we introduce a fast and low-cost method for capturing human hand demonstrations and converting the collected data into the ARCTIC format~\citep{ref12}, enabling different algorithms and models to perform data acquisition and training based on this framework. In Section~\ref{sec:dynamic_annealing}, we extract motion trajectories and contact information from demonstrations following the DexMachina framework~\citep{ref14} and construct motion imitation and contact-based auxiliary rewards. Building upon this, we further propose a contact-reward-based dynamic demonstration annealing mechanism, which enables policies to stably transition from demonstration-guided learning to autonomous reinforcement learning under low-cost demonstration conditions, thereby improving the success rate of articulated object manipulation across different dexterous hand embodiments.

\subsection{Dataset Construction}
\label{sec:data_collection}
We capture bimanual hand motions using a single RGB-D camera (Fig.~\ref{fig2}). Hands are first detected in 2D using MediaPipe~\citep{ref37}, and wrist 3D displacements are recovered using depth information. Let $(u,v)$ denote the wrist pixel coordinates and $z$ the depth, with camera intrinsics $(f_x, f_y, c_x, c_y)$. The wrist 3D position $(x,y,z)$ is computed as:
\begin{equation}
x = \frac{(u-c_x) z}{f_x}, \quad y = \frac{(v-c_y) z}{f_y}, \quad z = z.
\end{equation}
We then fit the MANO hand model~\citep{ref38} online to recover hand pose parameters $\theta$ and global wrist rotation $r$, by minimizing the keypoint reconstruction error:
\begin{equation}
\min_{\theta,r} \sum_{i=1}^{N} \omega_i \| J_i(\theta,r) - \hat{J}_i \|_2^2,
\end{equation}
where $J_i(\theta,r)$ denotes the predicted joint positions from MANO, $\hat{J}_i$ are detected keypoints, and $\omega_i$ are joint weights.

\begin{figure}[!htbp]
    \centering
    \includegraphics[width=\linewidth]{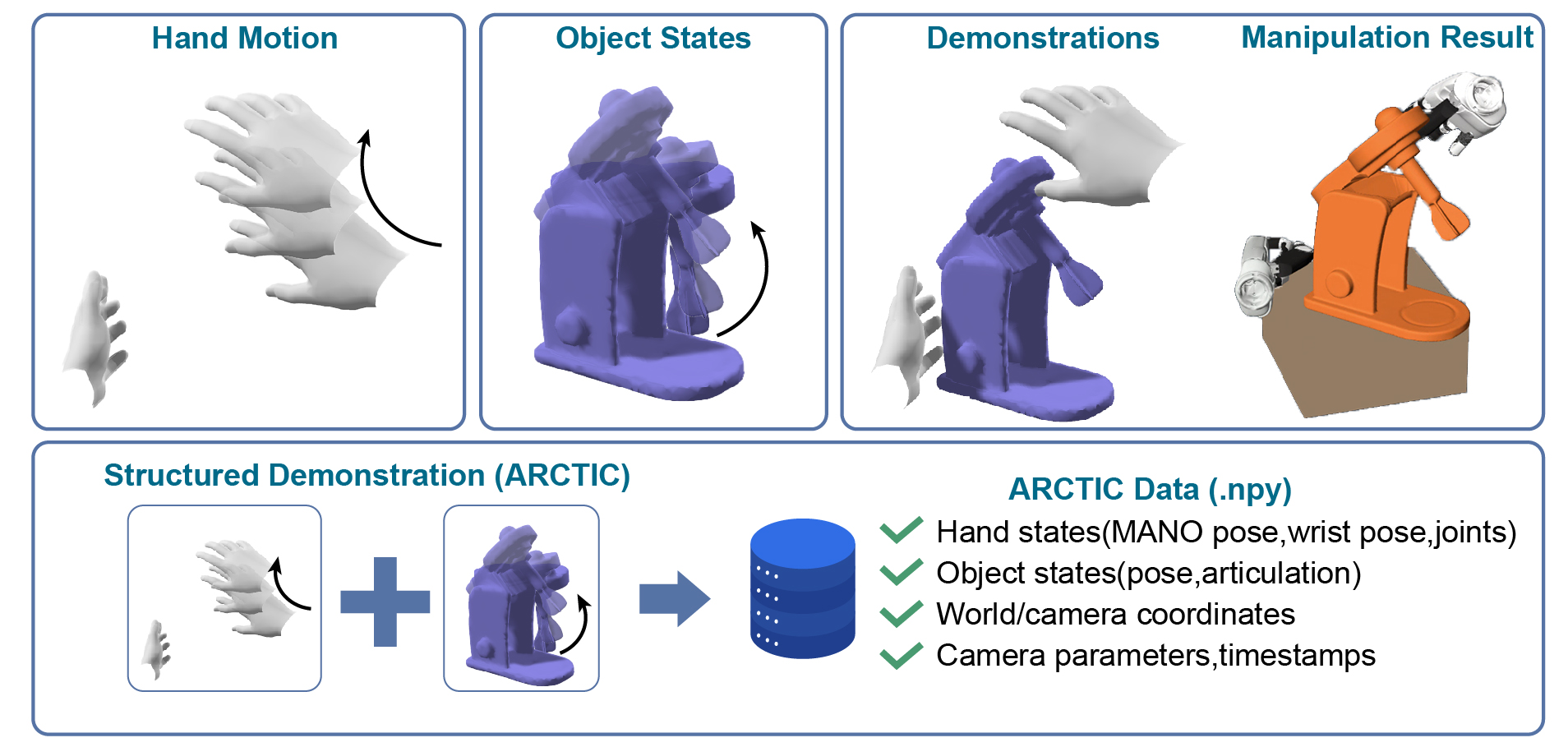}
    \caption{
    \textbf{Dataset construction process}.
    Predefined articulated object trajectories are first generated, after which the operator controls both hands to follow the object motion and produce feasible bimanual manipulation demonstrations under low-cost RGB-D capture. All interaction data are stored in ARCTIC format for downstream motion retargeting and policy learning.
    }
    \label{fig3}
\end{figure}

As shown in Fig.~\ref{fig3}, we adopt a trajectory-guided teleoperation strategy, where operators follow predefined object trajectories to produce feasible bimanual manipulation demonstrations under low-cost RGB-D perception. All interaction data are stored in ARCTIC format to ensure compatibility with existing tools and facilitate straightforward use by researchers familiar with ARCTIC. To further reduce high-frequency temporal noise introduced by RGB-D perception and online MANO fitting, we apply Gaussian smoothing along the temporal axis:
\begin{equation}
\hat{x}_t = \sum_{k=-r}^{r} g(k) \, x_{\text{clip}}(t+k,0,T-1),
\end{equation}
where $g(k)$ is a 1D Gaussian kernel. This improves temporal continuity of motion sequences.
\subsection{Contact-Reward-Based Dynamic Demonstration Annealing}
\label{sec:dynamic_annealing}

\begin{figure}[!htbp]
    \centering
    \includegraphics[width=\linewidth]{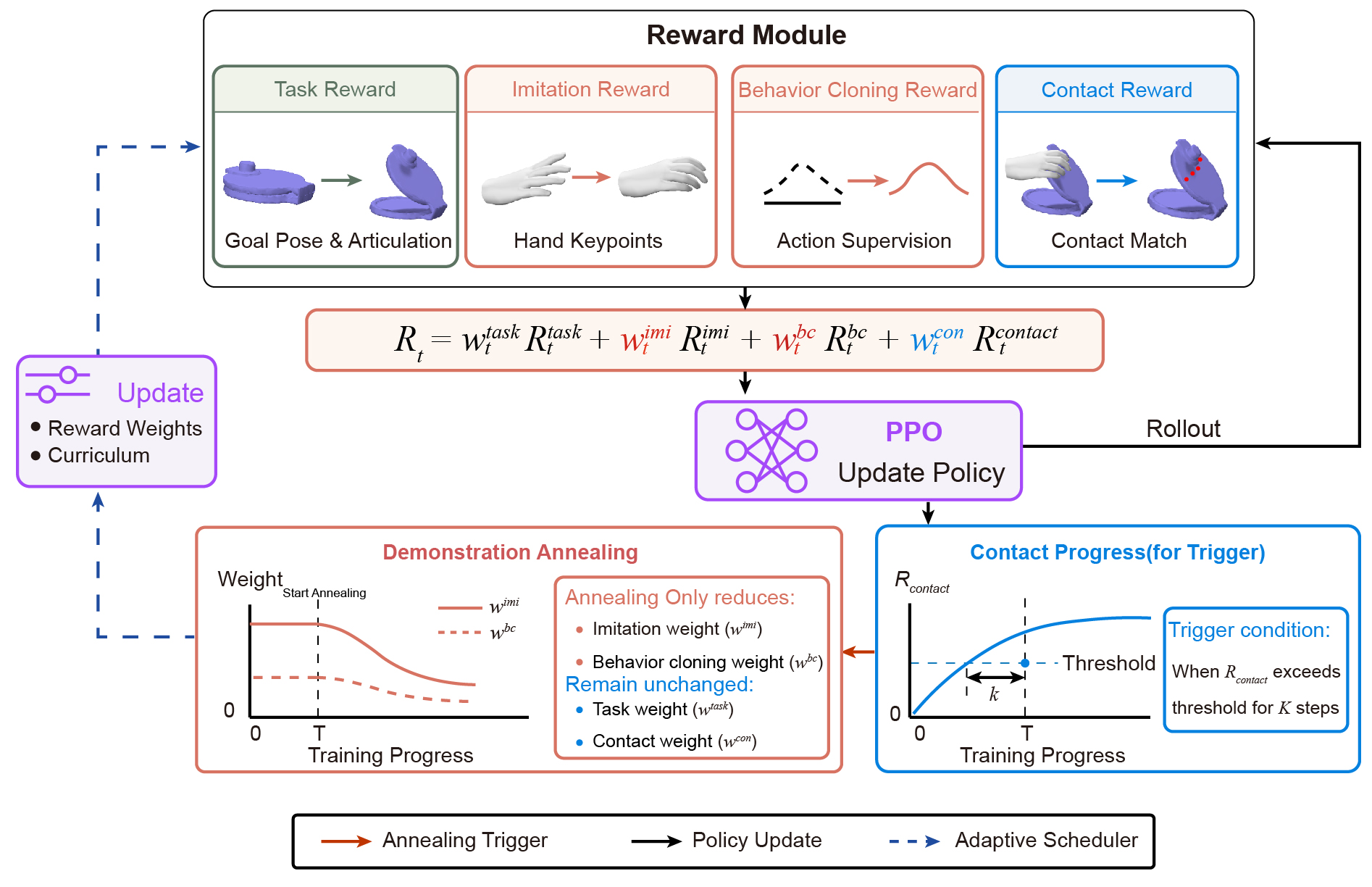}
    \caption{
    \textbf{PPO + Demo Annealing}. The policy is optimized using PPO with task, imitation, behavior cloning, and contact rewards. During training, contact progress is monitored to trigger adaptive annealing of imitation-related rewards, gradually reducing demonstration dependence while preserving task and contact objectives.
    }
    \label{fig4}
\end{figure}

During training, the object state sequences corresponding to human demonstrations are treated as target trajectories, and a bimanual dexterous hand policy is learned to drive the objects to follow these trajectories. As shown in Fig.~\ref{fig4}, we adopt PPO~\citep{ref39} as the base RL algorithm. The final reward is composed of task reward, demonstration-related reward, and contact reward:
\begin{equation}
r_t = w_{\text{task}} r_{\text{task}} + w_{\text{imi}} r_{\text{imi}} + w_{\text{bc}} r_{\text{bc}} + w_{\text{con}} r_{\text{con}},
\end{equation}
Here, $r_{\text{task}}$ measures the deviation between the current object state and the target trajectory. Let $d_{\text{pos}}$, $d_{\text{rot}}$, and $d_{\text{ang}}$ denote the position, rotation, and articulation-angle errors, respectively. The task reward is defined as:
\begin{equation}
r_{\text{task}}
=
\exp(-\beta_{\text{pos}} d_{\text{pos}})
\cdot
\exp(-\beta_{\text{rot}} d_{\text{rot}})
\cdot
\exp(-\beta_{\text{ang}} d_{\text{ang}}),
\end{equation}
where $\beta_{\text{pos}}$, $\beta_{\text{rot}}$, and $\beta_{\text{ang}}$ control the sensitivity of the position, rotation, and articulation-angle error terms, respectively. In addition to the task reward, we introduce demonstration-related rewards and a contact reward as auxiliary guidance. Specifically, $r_{\text{imi}}$ encourages the robot hand state to match the retargeted demonstration state, $r_{\text{bc}}$ regularizes the policy output toward retargeted demonstration commands, and $r_{\text{con}}$ encourages the policy to establish reasonable contact behaviors. 

Fixed demonstration weights may either overfit noisy demonstrations or increase exploration difficulty. To address this issue, we propose a contact-reward-based dynamic demonstration annealing mechanism, as illustrated in Fig.~\ref{fig4}. We use the progress of contact rewards as the trigger signal for dynamic demonstration annealing. Demonstration-related rewards are decayed only when the policy consistently achieves stable contact behaviors.

Let $w_{\text{imi}}^{(e)}$ and $w_{\text{bc}}^{(e)}$ denote the demonstration-dependent reward weights at epoch $e$, with lower bounds $w_{\text{imi}}^{\min}$ and $w_{\text{bc}}^{\min}$, and let $\gamma \in (0,1)$ be the annealing factor. When the annealing conditions are satisfied, the weights are updated as:
\begin{equation}
w_{\text{imi}}^{(e+1)} = \max(w_{\text{imi}}^{\min}, \gamma w_{\text{imi}}^{(e)}), \quad
w_{\text{bc}}^{(e+1)} = \max(w_{\text{bc}}^{\min}, \gamma w_{\text{bc}}^{(e)}).
\end{equation}
The key reward for annealing is $r_k$, which corresponds to the contact reward in our implementation. Let $l^{(e)}$ denote the average episode length at epoch $e$. 
Their averages over a sliding window of length $L$ are:
\begin{equation}
\bar{r}_k^{(e)} = \frac{1}{L} \sum_{i=e-L+1}^{e} r_k^{(i)}, \quad
\bar{l}^{(e)} = \frac{1}{L} \sum_{i=e-L+1}^{e} l^{(i)}.
\end{equation}
Dynamic demonstration annealing is executed only when:
\begin{equation}
e \ge e_{\text{wait}}, \quad
e - e_{\text{last}} \ge C, \quad
\bar{r}_k^{(e)} \ge \tau_k, \quad
\bar{l}^{(e)} \ge T_{\text{stable}}.
\end{equation}
Here, $e_{\text{wait}}$ denotes the minimum waiting epoch before annealing starts, $e_{\text{last}}$ is the epoch when the last annealing event occurred, $C$ is the cooling interval between two annealing events, $\tau_k$ is the threshold for the key reward, and $T_{\text{stable}}$ is the threshold for the sliding-window average episode length. 
%===========================================================================
\section{Experiments}
\label{sec:experiments}

This section evaluates the effectiveness of EaDex from three perspectives. First, we assess the overall performance of the proposed framework on cross-embodiment dexterous manipulation tasks using a custom low-cost demonstration dataset. Second, we perform ablation experiments by removing the dynamic demonstration annealing mechanism to analyze its effect on training stability and task success. Finally, we introduce dynamic demonstration annealing on the ARCTIC official dataset~\citep{ref12} and compare it with the original DexMachina~\citep{ref14} setup to examine whether the approach also improves policy learning under high-quality demonstration conditions.

\subsection{Experimental Setup}
\label{sec:exp_setup}

\textbf{Data Collection.} We evaluate EaDex under low-cost demonstration conditions using a custom human-hand interaction dataset. Only human demonstrations of three articulated objects are collected. All demonstrations are captured using the proposed low-cost RGB-D acquisition pipeline and organized in the ARCTIC format for downstream motion retargeting and policy training. The same set of human demonstration trajectories is then retargeted to three different bimanual dexterous hand embodiments, forming a total of $3 \times 3$ cross-embodiment dexterous manipulation tasks. This setup substantially reduces the cost of collecting demonstration data for multiple robotic hand configurations.

\textbf{Training.} All policies are trained in the Genesis physics simulation environment using PPO as the base reinforcement learning algorithm. Across all hand embodiments and tasks, the policy operates in a unified state-based observation space and simultaneously controls both hands for coordinated manipulation. We use three random seeds for experiments on our custom dataset and five random seeds for experiments on the ARCTIC dataset.

\textbf{Evaluation Protocol.} For the custom low-cost demonstration dataset, each articulated object is placed on a platform of size $0.2\,\mathrm{m} \times 0.2\,\mathrm{m} \times 0.1\,\mathrm{m}$. A rollout is considered successful if the object remains on the platform throughout the episode and the final articulation angle exceeds $45^\circ$. This protocol accounts for both task completion and operational stability, providing a comprehensive measure of the policy's manipulation capability. For the ARCTIC dataset, evaluation is conducted using the ADD-AUC3 metric from DexMachina to ensure consistency with prior work.

\textbf{System Efficiency.} During data collection, we use a single Intel RealSense D435i RGB-D camera and an Intel i9-10900KF CPU to capture human-hand interaction sequences in real time. Policy training is performed in the Genesis simulator on a single NVIDIA RTX 3090 GPU. Compared with traditional motion-capture systems, complex teleoperation setups, or large-scale GPU training infrastructure, the proposed framework substantially reduces the hardware and deployment cost of dexterous manipulation learning. Under the current setup, the complete pipeline from demonstration collection to successful policy learning can be completed within approximately 1 hour for some tasks.

\subsection{Results on Low-Cost Demonstrations}
\label{sec:low_cost_results}

\begin{figure}[!htbp]
    \centering
    \includegraphics[width=\linewidth]{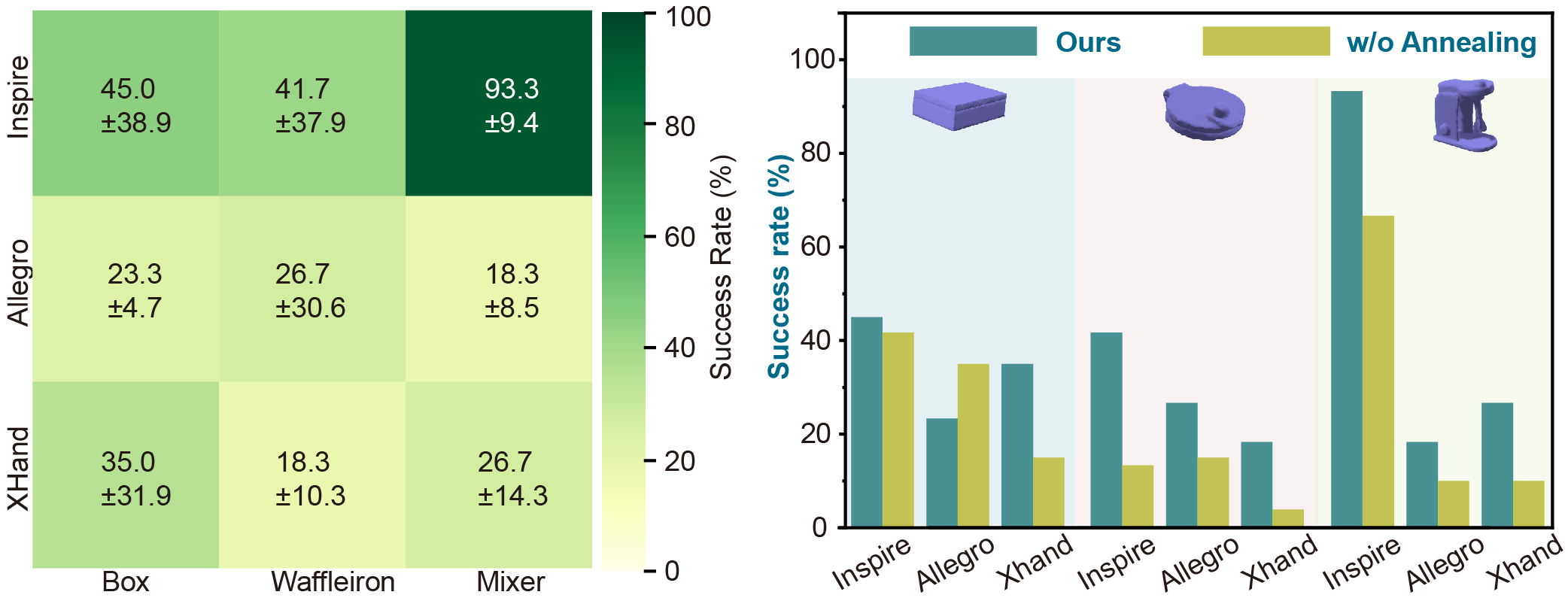}
    \caption{
    \textbf{Experimental Results for Cross-Embodiment Manipulation}.
    Experiments are conducted on our custom dataset.
    \textbf{Left}: Task success rates across embodiments and tasks with demonstration annealing.
    \textbf{Right}: Ablation with and without annealing.
    }
    \label{fig5}
\end{figure}

We first evaluate EaDex on the custom low-cost demonstration dataset. The experiment covers three bimanual dexterous hand embodiments and three articulated objects, resulting in a total of nine cross-embodiment manipulation tasks. As shown in the heatmap of Fig.~\ref{fig5}, EaDex achieves an average success rate of 36.5\%, with some tasks reaching a maximum success rate of 93.3\%. Overall, EaDex is able to perform multiple articulated object manipulations under low-cost RGB-D demonstration conditions, indicating that the proposed data acquisition and policy learning framework supports cross-embodiment dexterous manipulation.

Examining task-specific results, success rates vary across different hand-object combinations. Some tasks achieve high success rates, showing that low-cost demonstrations can still provide effective motion structure and contact priors. More challenging tasks show lower success rates mainly due to object geometry; in particular, thinner objects make it harder for dexterous hands to establish stable contacts, resulting in reduced manipulation performance. 
\enlargethispage{2\baselineskip}
\subsection{Ablation Study on Dynamic Demonstration Annealing}
\label{sec:ablation}
To evaluate the effectiveness of the contact-reward-based dynamic demonstration annealing mechanism, we perform ablation experiments on the custom low-cost demonstration dataset. All experiments use the same demonstration data, reward design, training budget, and PPO configuration; the only difference is whether dynamic demonstration annealing is applied.

Specifically, in the \textit{w/o Annealing} setting, the weights of the imitation reward and behavior cloning reward are kept fixed throughout training. In the \textit{Ours} setting, these demonstration-related reward weights are dynamically decayed according to the policy's contact learning progress. The histogram in Fig.~\ref{fig5} presents the success rates of both settings across the nine cross-embodiment dexterous manipulation tasks.

The results show that dynamic demonstration annealing significantly improves task success rates under low-cost demonstration conditions. Compared to the fixed-weight setting, our method increases the average success rate from 23.5\% to 36.5\%, a relative improvement of 55.3\%. This demonstrates that fixed demonstration weights are prone to being affected by noise in low-cost demonstrations, whereas dynamic annealing retains demonstration guidance during early training and gradually releases it as the policy achieves stable contact behaviors, thereby improving task success rates.

\subsection{Validation on the ARCTIC Dataset}
\label{sec:arctic_validation}

To further examine the applicability of dynamic demonstration annealing, we conduct additional experiments on the official ARCTIC dataset, which provides higher-quality human-hand interaction demonstrations than our custom low-cost dataset. This experiment assesses whether dynamic demonstration annealing can also improve policy learning for complex contact tasks under high-quality demonstration conditions.

We select the mixer, ketchup, and waffleiron tasks for the Ability Hand from DexMachina as evaluation targets, and incorporate the proposed dynamic demonstration annealing mechanism into its original training setup. To ensure a fair comparison, we keep the task configuration, training procedure, and ADD-AUC3 evaluation protocol as consistent as possible. The baseline results are taken from the original DexMachina paper.

As shown in Table~1, dynamic demonstration annealing improves policy performance on several complex contact tasks, indicating that the mechanism is effective not only for low-cost demonstrations but can also serve as a general strategy for scheduling demonstration constraints to improve task success rates in demonstration-guided reinforcement learning.

\begin{table}[htbp]
\centering
\caption{ADD-AUC3 success rates (\%) for Ability Hand tasks on ARCTIC dataset.}
\begin{tabular}{lccc}
\toprule
Settings & Ketchup & Waffleiron & Mixer \\
\midrule
DexMachina (Unannealing) & 9.0 $\pm$ 0.6 & 9.1 $\pm$ 0.7 & 28.1 $\pm$ 7.4 \\
Ours (Annealing) & \textbf{57.91 $\pm$ 28.12} & \textbf{23.01 $\pm$ 0.65} & \textbf{35.14 $\pm$ 4.02} \\
\bottomrule
\end{tabular}
\label{tab:arctic_results}
\end{table}
%===============================================================================

\section{Conclusion and Limitations}
\label{sec:conclusion}

\textbf{Limitations.} 
Our method uses a single RGB-D camera to capture hand motions. When gestures are occluded, keypoint detection may be incomplete. To mitigate this issue, we adopt a reference pose with local offsets when controlling the wrist, ensuring the palm faces the camera as much as possible, which reduces occlusion. However, this approach can slightly limit the dexterity of wrist motions. Future work may address this limitation by using additional cameras and multi-view fusion techniques~\citep{ref40}.

\textbf{Conclusion.} 
In this work, we propose a low-cost demonstration data acquisition method and apply it to guide reinforcement learning policy training. Experimental results demonstrate that our approach significantly reduces the cost of data collection and policy training, accelerates learning, and enables bimanual dexterous manipulation of various articulated objects. This provides a practical and cost-effective solution for real-world dexterous manipulation learning tasks.

\bibliography{example}  % .bib

\end{document}